\documentclass[nocopyrightspace]{sig-alternate}

\def\thepage{\arabic{page}} 

\usepackage{pifont}
\usepackage{times-lite,color}
\usepackage{paralist}
\usepackage{semantic}
\usepackage{graphicx}
\usepackage{xspace}
\usepackage[caption=false]{subfig}
\usepackage{xcolor}
\usepackage{listings}
\usepackage[T1]{fontenc}
\usepackage{minibox}

\usepackage{sfmath}
\usepackage{balance}

\IfFileExists{.usepdfcomments}
{\usepackage[icon=Comment,color=notecolor,hoffset=-5pt,voffset=8pt]{pdfcomment}
\def\usepdfcomment{}}

\usepackage{ttquot}

\definecolor{javapurple}{rgb}{0.5,0,0.35}
\definecolor{javagreen}{rgb}{0,0.6,0}

\renewcommand{\sp}{\ | \ }

\newcommand{\textred}[1]{\textcolor{red}{#1}}
\ifx\noeditingmarks\undefined%
\ifx\usepdfcomment\undefined%
\newcommand{\pgwrapper}[4]{%
\def\argone{#1}%
\def\none{None}%
\ifx\argone\none%
\textred{#2: #3}%
\else%
\textred{#1 [#2: #3]}%
\fi}%
\else%
\definecolor{notecolor}{rgb}{1.0,1.0,0.7}%
\definecolor{srmcolor}{rgb}{1.0,1.0,0.7}%
\definecolor{aslcolor}{rgb}{1.0,0.8,0.7}%
\definecolor{accolor}{rgb}{0.7,1.0,1.0}%
\newcommand\pgwrapper[4]{%
\def\argone{#1}%
\def\none{None}%
\ifx\argone\none%
\pdfcomment[author=#2,color=#4]{#3}%
\else%
\pdfmarkupcomment[author=#2,color=#4]{#1}{}%
\pdfmargincomment[author=#2]{#3}\fi}%
\fi
\else
\newcommand{\pgwrapper}[4]{#1}
\fi

\newcommand{\figlabel}[1]{\label{f:#1}}
\newcommand{\seclabel}[1]{\label{s:#1}}
\newcommand{\secref}[1]{Sec.~\ref{s:#1}}  
\newcommand{\figref}[1]{Fig.~\ref{f:#1}}     
\newcommand{\Figref}[1]{Figure~\ref{f:#1}}   

\usepackage{hyperref}

  \renewenvironment{thebibliography}[1]{%
    \begin{oldthebibliography}{#1}%
      \setlength{\parskip}{0ex}%
      \setlength{\itemsep}{0ex}%
  }%
  {%
    \end{oldthebibliography}%
  }

\newcommand{\C}[1]{{\small\lstinline!#1!}}

\renewcommand{\b}[1]{\ensuremath{{\sf{#1}}}}

\renewcommand{\min}{\b{min}}

\renewcommand{\sum}{\b{sum}}

\newcommand{\AND}{\wedge}
\newcommand{\OR}{\vee}

\newcommand{\Sk}{\textsc{Sketch}\xspace}

\newcommand{\cmark}{\ding{51}}%
\newcommand{\xmark}{\ding{55}}%

\begin{document}

\title{Using Program Synthesis for Social Recommendations}

\numberofauthors{1}
\author
{
  \alignauthor
  Alvin Cheung \quad Armando Solar-Lezama \quad Samuel Madden\\
  \vspace{0.1in}
  \affaddr{MIT CSAIL}\\
  \vspace{0.1in}
  \email{\{akcheung, asolar, madden\}@csail.mit.edu}
}

\maketitle

\begin{abstract}

This paper presents a new approach to select events of interest to a user in a
social media setting where events are generated by the activities of the user's
friends through their mobile devices.  We argue that given the unique
requirements of the social media
setting, the problem is best viewed as an inductive learning problem, where the
goal is to first generalize from the users' expressed ``likes'' and ``dislikes''
of specific events, then to produce a program that can be manipulated by the
system and distributed to the collection devices to collect only data of
interest.

The key contribution of this paper is a new algorithm that combines existing
machine learning techniques with new program synthesis technology to learn
users' preferences. We show that when compared with the more standard approaches,
our new algorithm provides up to order-of-magnitude reductions in model
training time, and significantly higher prediction accuracies for our target
application. The approach also improves on standard machine learning techniques
in that it produces clear programs that can be manipulated to optimize data
collection and filtering. 
\footnote{{\small A shorter version of this paper appeared in CIKM'12.}}

\end{abstract}



\section{Introduction}
\label{s:introduction}

At a high level, the problem of selecting events or updates of interest to a
user in a social media setting appears similar to recommendation problems in
other environments, such as offering book or movie recommendations on Amazon and
Netflix. In each of these, a user's previously expressed preferences are used to
infer new items of interest; every time the user interacts with the site, the
system builds a more accurate picture of what she likes and dislikes and uses it
to improve recommendations. Social media, however, poses some unique challenges
which demand a different approach from the standard {\it collaborative
filtering}, where other users' preferences are used to infer
about what the user will like~\cite{neighborsNetflix, sideInformation}.

To illustrate some of the new challenges that recommendation systems face in
this domain, we focus on an application called LifeJoin~\cite{lifeJoin}.  We
designed this application to model the future of social networking, where a
person's profile is continuously updated (modulo a privacy filter) by an
automatically generated event stream from the user's mobile devices, including
her location and activities (e.g., running, sitting on a bus, in a meeting,
etc).
The system also attempts to discover interesting co-occurrences in friends'
event streams, such as a meeting of two of the user's friends in a
nearby pub.  In order to deal with the data deluge, the system gives the user
the ability to ``like'' and ``dislike'' both individual and combinations of
events. LifeJoin uses the expressed likes and dislikes to infer what kinds of
events are of interest to the user, which can then be used to auto-populate the
user's newsfeed or notify her of interesting nearby social events. 
Collecting all sorts of events through a mobile device can 
consume a lot of energy \cite{powertutor}, 
so LifeJoin uses the
inferred user's interest to drive subsequent event acquisition. For instance, if
LifeJoin infers that Mary's friends are only interested in the places she
goes for a jog, then the system will save power on Mary's device by turning off
data collection when she is not jogging. Our initial experiments have shown that
implementing the data collection scheme in the scenario above can extend the 
phone battery life by up to 40\% \cite{lifeJoin}. 
Thus, the more accurate we can detect the users'
real interests, the more energy we can save in data collection as compared to 
a scheme that collects all data under all circumstances.

More specifically, inferring interests in LifeJoin poses 
four unique challenges: 
\begin{asparaenum}
\item \textbf{Decomposable Models:} 
 For applications such
as LifeJoin, models must be decomposable into simple classifiers that
can be pushed down to the individual devices to drive event acquisition.
One simple way to ensure a model is decomposable is to
limit it to only contain boolean combinations of simple predicates
over the input features, which can be decomposed in a straightforward
way to indicate the required data from phones. Such models are
also useful because they allow users to give explicit feedback about
whether the system actually understands their true interests, and
to manually tune the models to better suit their preferences as discussed
in \cite{neighborsNetflix}. 
By contrast, many existing preference learning algorithms produce black box
classifiers that are difficult to decompose and understand.

\item \textbf{Active Learning:} Given the large number of incoming events, and
the large number of ways in which they could be combined, it is unreasonable to
ask the user to rate any meaningful fraction of them. Thus, the learner needs
to intelligently choose a subset of incoming events that can most improve the
current model.
In addition, the domain of users mentioned in the
incoming events can also change over time as the user's friends network changes. 

\item \textbf{Noisy, Skewed Data:} 
Since the ratings are produced by humans, they are bound to contain occasional
errors.  Users also change their interests over time, so the same event might be
given different ratings depending on when it was shown to the user.  At the same
time, each user's definition of ``interesting'' is different, so it is difficult
to make generalizations about the statistical properties such as the anticipated
degree of skew in users preferences. In fact, this is currently an active research
topic on its own \cite{interestDrift}.

\item \textbf{Personalized Events:} Unlike typical recommendation systems such
as those for books, movies, or online ads, where all users rate a common set of
items, the events in LifeJoin tend to be highly personalized.  For instance, a
user might like an event because it involves her best friend Peter, but the same
event would be totally meaningless if is shown to another user who does not know
Peter.  
Thus, we believe it is easier to learn a model for each user individually 
(i.e., event is of interest if it involves Peter, without needing to know the
relationship between Peter and the user) rather
than trying to discover the relationships between users and design a model that
is applicable to all.
\end{asparaenum}

These requirements preclude the use of collaborative filtering (CF) techniques
which have been successful in other recommendation systems---such as building
neighborhood or latent factor models to predict user ratings. In particular,
these techniques tend to generate models that cannot be used to drive data
acquisition and generate an explainable model to the user to solicit 
further feedback (req 1).  For instance, a neighborhood model-based approach might
attribute a new rating
based on a set of previously rated events that are deemed similar, but it is
unclear how the system can easily generalize from the set of similar events to
determine what new events to collect.  Furthermore, CF techniques require a
similarity measure between users or events.  It is unclear how that can be done
in a setting where events are highly personalized to a small set of users (req
4); this is an active research topic~\cite{linkPrediction,socialTwist}, 
and the proposed solutions require explicitly modeling all 
 social relationships between users, rather than simply learning a separate
 model for each user individually, which does not require discovering the 
 relationships among the users.

We avoid the above issues by viewing the problem as an inductive learning
problem with an active learning component: given a set of labeled examples, the
goal is to learn a set of rules that represents an individual 
user's preferences, and to
choose new events for the user to rate.  Unfortunately,
standard inductive learning algorithms such as those based on entropy measures
(e.g., decision trees and inductive logic programming tools) are known have
issues with skewed data (req 3) \cite{imbalancedData}, it is not clear how active
learning can be applied,
and they also do not provide good generalization guarantees when compared to
statistical-based learners such as support vector machines (SVM).  

Recently, the programming languages community has been exploring
inductive learning problems in the context of software synthesis in
programming-by-example systems \cite{excelSynthesis}, 
where the goal is to infer a program from a set of sample behaviors.
Unfortunately, the learning problem in Lifejoin is different enough that none
of the previous techniques from this community can be applied out of the box.
In particular, the active learning problem has not been sufficiently addressed
by previous research from this community. Nevertheless, these techniques
provide a new set of tools that can be leveraged to attack the problem.

In this paper, we present a new algorithm to infer users' interests; the
algorithm combines new techniques in program synthesis with more traditional
machine learning approaches to satisfy the unique requirements illustrated by
the LifeJoin application. Specifically, we make the following contributions:

\begin{asparaenum}

\item We show that both the classical machine learning approach and an
  approach based purely on program synthesis do not adequately address
  this problem.  

\item We describe a hybrid approach that employs program
  synthesis to generate a number of classifying functions,
  and subsequently asks an SVM to assign weights to the features in each
  generated functions.
  We show that, when compared to pure machine learning or synthesis
  approaches, this hybrid technique 
  takes up to an order of magnitude less
  time to encode the training data into a feature space
  representation, and 
  improves upon traditional learning algorithms by
  30\% in overall classification accuracy.
\item We show that we can use a program synthesizer to produce  
   more decomposable and human-understandable models than those generated by
  traditional machine learning techniques, and provide empirical
  evidence that the generated models are comparable to the original
  intentions that the user has in her mind.

\end{asparaenum}

We have implemented the learning technique in the context of the LifeJoin
application. However, we believe that our approach is applicable to other 
social networking applications as well, where large amounts of data
are collected from users, and labels provided by users contain errors or
interest drifts.
In the next section we give an overview of the various steps in the
learning task in LifeJoin, and illustrate our approach with
an example.

\section{Overview of the approach}
\label{s:overview}

\begin{figure*}
\begin{minipage}{2.15in}

\begin{tabular}{|c|c|c||c|}
\hline
user & location & time & preference \\
\hline
Joe & Office & 10am & \xmark \\
Bill & Home & 3pm & \xmark \\
Joe & Office & 11pm & \cmark \\
Joe & Bar & 6am & \cmark \\
\hline
\end{tabular}
\\

\end{minipage}
\quad \vrule \quad
\begin{minipage}{4in}
\begin{tabular}{c}
\\
 Each line below denotes a potential classifier \\
 \hline
 \\
 {\small
 $(User=Joe) \wedge (location=Office \vee location=Bar) \wedge (time < 7am \vee time > 10pm)$ }\\
 {\small $(User \neq Bill) \wedge (time > 10pm \vee  location=Bar)$ }\\
 {\small $(User=Joe) \wedge (time < 9am \vee time>11am) $ }
\\
\end{tabular}
\end{minipage}
\caption{Learning example with labeled data (left) and candidate classifiers that
are consistent with the labeled data (right)} 
\figlabel{runningex}
\vspace{-0.2in}
\end{figure*}

In this section, we illustrate the recommendation problem with a concrete
example and present an outline of our solution. To frame the problem, consider
the LifeJoin event stream, which contains large numbers of events about the
activities of a user's friends and family. Out of this event stream, suppose
that the user is interested in events where her friend Joe is away from home
either late at night or early in the morning: 
\begin{small}
\begin{eqnarray*}
(user = Joe) \,\AND\, (location \ne Home) \,\AND\, (time < 9am \OR time > 9pm)
\end{eqnarray*}
\end{small}
The goal of the system is to infer this interest function based on events the
user rates as having liked or disliked. We want the algorithm to produce its
interest function in the form of a predicate like the one above because that
helps ensure the decomposability described earlier. When the interest function
is expressed in this form, it can be easily manipulated and decomposed
into predicates that can be pushed down to individual users' phones to optimize
the data acquisition process as described.  Such expressions are also
comprehensible by 
users, and can be manually adjusted to tune the results the user sees.
We are not aware of any statistically-based methods (such as CF or SVM) 
that can directly generate models like these.

In the absence of additional information about the expected distribution of the
events, the most na\"ive approach to finding an interest function is to
exhaustively explore the space of all possible predicates of the desired form
until a set of predicates is found that matches all the previously labeled
events. The most obvious problem with such an approach is that the space of
possible predicates is enormous---on the order of $10^{40}$ in some of our
experiments. However, as we will describe in \secref{plApproach}, new
technology from the field of combinatorial synthesis~\cite{sketch06} can find a
matching interest function in this space in a few seconds. For example,
\figref{runningex} shows a sample of labeled data and a few interest functions
that were found this way to match the data.

A deeper problem with the na\"ive approach is that predicates found this way
cannot be expected to have much generalization power---that is, they are
unlikely to correctly classify as yet unseen data items. Individually, they
will also not be of much use in optimally determining the next data element to
present to the user for labeling. To address this problem, we rely on the idea
of boosting \cite{boosting}. After the combinatorial synthesis algorithm has
found $K$ interest functions $f_i$, each of these functions can be treated as a
weak base learner, and the group forms an ensemble.

The standard way of forming the ensemble is to learn a linear function
$F(e)=\Sigma w_i\cdot f_i(e)$, where an event is classified as interesting if
$F(e)>0$. The ensemble allows us to follow a standard approach for active
learning, namely, to select those events that are closest to the boundary where
$F(e)=0$~\cite{svmActiveLearning}. Normally, the weights $w_i$ are selected
based on the training data, but in our case, since all the functions $f_i$ were
selected to agree on all the training events, that leads to all functions
having equal weight. That means that the ensemble reduces to a majority vote,
and the active learning strategy reduces to selecting the event that causes the
maximum level of disagreement among all the candidate interest functions. 
We refer to this pure synthesis based algorithm as the ``ensemble'' approach.
As we
will see in \secref{experiments}, such an approach already outperforms many
standard learning techniques, but we can do better. 

When defining the space of candidate interest functions, we require the
functions to be in disjunctive normal form. This means that every function
$f_i$ can be seen as a disjunction of individual predicates $p_{i,j}$. We exploit
this structure when building the ensemble; instead of an ensemble $F(e)=\Sigma
w_i\cdot f_i(e)$, we build an ensemble of the form $F'(e)=\Sigma w_{i,j} \cdot
p_{i,j}(e)$. Finding weights for each predicate is no longer trivial. We
use an SVM to find a set of weights for the function, which has the additional
benefit that the weights will be set in such a way that the resulting classifier
will be maximum-margin one.  As we will see in
\secref{experiments}, defining the ensemble in this way significantly improves
active learning, and we call this combination of program synthesis and machine
learning techniques the ``hybrid'' approach.

As our experiments show, this approach also copes gracefully with errors in the
training data. 
We can improve its handling of
errors by selecting each of the $f_i$ functions to match only a randomly chosen
subset of the data. If the rate of errors is low, this ensures that at least
some of the $f_i$ will be selected to match only uncorrupted data.

One issue that still has to be addressed is that the SVM may find fractional
values for the weights, so the function $F'(e)$ will no longer be a well-formed
boolean predicate. Once again, we use combinatorial synthesis technology to
find a well-formed predicate $P(e)$ that is closest to the linear function
$F'(e)$.   Such predicates have the decomposability property we desire.

Now that we have described our basic approach and problem setup, we describe the
synthesis technology we use to solve the problem in more detail in
\secref{plApproach}, as well as the details of our hybrid approach in
\secref{hybridApproach}. \secref{experiments} presents our experiments on a
synthetic data set derived from the LifeJoin scenario, showing substantial
performance gains for the hybrid approach. Finally, we discuss related work in
\secref{relatedWork}, and conclude in \secref{conclusions}.

\section{Constraint Based Synthesis}
\label{s:plApproach}

In recent years, there has been a lot of interest in the programming languages
community around constraint-based approaches to program
synthesis~\cite{sketch06, sketch08, oracleGuidedSynthesis,
pathBasedSynthesis}. At a high-level, this technology
provides an efficient mechanism to search a space of candidate programs for one
whose behavior satisfies a given specification. 

The synthesis problem can be seen as a generalization of traditional curve
fitting, where a space of possible curves---say, the space of all polynomials
of degree less than $k$---is explored in search of one that satisfies a given
set of requirements. Modern synthesis systems go several steps beyond simple
curve-fitting by providing rich languages for describing requirements and
spaces of candidate programs. The search for a correct solution in this space
is performed symbolically; i.e., the space of candidate programs is described
through a set of equations which are solved through a combination of inductive
and deductive methods by a specialized solver. LifeJoin uses a synthesis system
called \Sk{} \cite{sketch06}. In the
reminder of this section, we give a brief overview of Sketch, and show how our
system uses it to generate candidate solutions to the problem.

\subsection{The Sketch Synthesis System at a Glance}

Sketch extends a simple procedural language --- think C or Pascal --- with new
constructs that allow users to write programs with \emph{holes}, i.e., missing
expressions that must be completed by the synthesizer. The language allows
programmers to use recursive definitions to describe the space of expressions
that can be used to fill a hole. For example, consider the following program: 
\begin{center}
\begin{tabular}{c}
\begin{small}
\begin{lstlisting}
int foo(int x, int y) { return expr(x,y); }

generator int expr(int x, int y) {
  return {| ?? | x | y | expr(x,y) + expr(x,y) |};
}
\end{lstlisting}
\end{small}
\end{tabular}
\end{center}
The program defines an expression \C{expr} to be either a constant, a variable
\C{x} or \C{y}, or a sum of similar sub-expressions. In essence, the generator
defines a grammar for the possible expressions that can be returned.

Given the grammar, the user can constrain the behavior of the desired 
expression by writing test
harnesses. For example, the test harness below ensures that the value returned
by the function is greater than twice the first parameter when the second
parameter is greater than zero, and ensures also that when 
\C{x=5} and \C{y=8} the function produces \C{10}. 
\begin{center}
\begin{tabular}{c}
\begin{small}
\begin{lstlisting}
harness void main(int px, int py) {
  if (py > 0) { assert foo(px, py) > 2*px; }
  assert foo(5, 8) == 12;
}
\end{lstlisting}
\end{small}
\end{tabular}
\end{center}
Given such a program as input, the Sketch system discovers that a 
plausible solution is for \C{foo} to return \C{x+x+2}. 

To understand how the technology works, consider the generator \C{expr} which
describes a set of possible expressions. One way to understand this generator
is that every time it is called, the system has to make a choice about what to
return. In order to turn \C{foo} into a concrete piece of code, the system
needs a strategy to make those choices; i.e., it needs to find a recipe for how
to make the choices in \C{expr} to ensure that the correct answer is produced
every time. Sketch encodes such a recipe as a vector of bits $\hat{c}$, so the
test harness can be seen as taking $\hat{c}$ as an additional parameter. The
goal of finding a strategy that works every time reduces to finding a value of
$\hat{c}$ that satisfies an equation of the form \[
	\forall px, py. P_{main}(px, py, \hat{c})
\]
The predicate $P_{main}$ is derived from the test harness \C{main}
automatically by a compiler, and is true if the strategy $\hat{c}$ causes the
function to pass the assertions when run with inputs $px$ and $py$. Sketch
translates the equation above into a series of boolean satisfiability problems.
Unlike traditional inductive learners such as decision trees (which gives 
poor results as discussed in \secref{introduction} 
and \secref{experiments}),
the core algorithm in Sketch works by forming an initial
hypothesis about the solution, then iteratively finds instances from the 
harness that fails the hypothesis and incorporates them into the hypothesis itself.
The process repeats until the harness is satisfied.
In practice, this tends to be quite fast in 
in terms of solution generation time as our experiments show, and the 
details of the algorithm are described in \cite{sketch06}. The algorithm itself
is NP-complete as it uses a SAT solver as the backend. However, in practice 
many of the problems, such as those in LifeJoin,
can be solved in very short time as our experiments show.

\subsection{Encoding the Space of Interests using Sketch}

Given the above, we now discuss how we use Sketch to aid in feature selection
in LifeJoin.
One of the problems with feature selection is that there is an exponentially
large space of possible features, so analyzing them one by one to identify
those that better predict the labels in the training data is prohibitively
expensive. By contrast, constraint-based synthesis allows us to represent the
entire space of possible interest functions as a compact sketch that uses a
grammar to describe the space of all possible solutions to the classification
problem.

As mentioned in \secref{overview}, we would like to generate interest functions
that select data elements from the stream of events collected from users' phones
by returning a boolean value given events from the event streams.
To generate the appropriate interest function, we encode its grammar using
Sketch similar to that of the example above.
LifeJoin currently collects two streams of events from users' phones: a stream
that describes a user's activity (walking, running, etc), and another that
describes a user's location. Both event streams come with timestamps that 
describe the start and end time of each event along with the user involved
and. Given that, we encode the space
of interest functions using a grammar with predicates from the two 
event streams, as shown in \figref{interestLanguage}. Each interest function
consists of a disjunction of interests and returns a boolean value.
Each interest takes in an activity and a location event, and consists of a 
conjunction of event predicates. Each event predicate is 
either one that restricts the set of events from either event stream, or is
a join predicate that links events from both data streams, for instance 
the user from the location event has to be the same as the user from the 
activity event. As an example, a user who is interested in events about Peter
running along the Charles River can be represented with the interest function: 

\begin{small}
\begin{eqnarray*}
a.user = Peter \AND a.activity = running \AND \\
l.location = Charles~River \AND a.user = l.user 
\end{eqnarray*}
\end{small}

We formulated our grammar based on initial user studies, and 
further predicates (such as average duration of events) can be incorporated
as needed.
In our experiments we also bound the maximum number of disjuncts and conjuncts
allowed in the interest functions and interests
during synthesis, along with the set of users, activities, and locations.

\begin{figure}
\begin{small}
\centering
{\jot=1pt
\begin{eqnarray*}
f(a,l) \in \b{interest~function} &::=& \bigvee_k~i_k(a,l) \\
i(a,l) \in \b{interest} &::=& \bigwedge_k~\left(ap(a) \sp lp(l) \sp jp(a,l)\right) \\
\\
a \in \b{activity} &::=& \{ user, activity, start, end \} \\
l \in \b{location} &::=& \{ user, location, start, end \} \\
ap(a) \in \b{activity~pred} &::=& a.user ~op~ \{ Users \} \\
                   &\sp& a.activity ~op~ \{ Activities \} \\
                   &\sp& a.start ~op~ N \sp a.end ~op~ N \\
                   &\sp& (a.end - a.start) ~op~ N \\
lp(l) \in \b{location~pred} &::=& l.user ~op~ \{ Users \} \\
                   &\sp& l.location ~op~ \{ Locations \} \\
                   &\sp& l.start ~op~ N \sp l.end ~op~ N \\
                   &\sp& (l.end - l.start) ~op~ N \\
jp(a,l) \in \b{join~pred} &::=& a.user ~op~ l.user 
                   \sp a.start ~op~ l.start \\
                   &\sp& a.end ~op~ l.end 
                   \sp a.start ~op~ l.end \\
                   &\sp& a.end ~op~ l.start \\
                   &\sp& (a.end - a.start) ~op~ (l.end - l.start)
\end{eqnarray*}
\vspace{-0.2in}
}
\end{small}
\caption{Grammar of Interests}
\vspace{-0.1in}
\figlabel{interestLanguage}
\end{figure}

Following the example above, with
the grammar for interest functions we use the previously-labeled events
from the user as the harness. We then ask the Sketch system to generate an interest
function that satisfies the labels on the training events. And each function 
generated becomes a weak base learner as discussed in \secref{overview}.

With this in mind, we next discuss how the weak base learners are combined in
the ensemble and hybrid approaches.

\section{The hybrid approach}
\seclabel{hybridApproach}

In this section we discuss in detail our ensemble and hybrid 
approaches, and provide insights
into why the hybrid one performs better than the ensemble one.

\subsection{Hybrid Algorithm}

As mentioned in \secref{overview}, our classifier works by first generating a
number of functions that are capable of fully explaining the training data.
Unfortunately, the ensemble approach does not provide any generalization 
guarantees. However, as we later pointed out in the same section, 
we can instead break the weak learners into their predicate constituents and
treat them as base features, and then use them as features to train a
SVM classifier. Classification is then done using the SVM, 
with events classified as interesting if it returns a value 
$\ge 0$, and is not interesting otherwise.  \Figref{hybridAlgorithm} outlines this
hybrid algorithm in pseudocode form.

\begin{figure}
\lstset{numbers=left, firstnumber=1, numberstyle = \tiny\color{gray}, 
        deletekeywords = {harness}}
\centering
\begin{tabular}{c}
\begin{small}
\begin{lstlisting}
learnModel (posEs, negEs) { @\label{lst:learnModel1}@
  (posTrainEs, negTrainEs) = subsample(posEs, negEs); @\label{lst:subsample}@
  baseFns = callSketch(harness, posTrainEs, negTrainEs);
  preds = extractPredicates(baseFns); @\label{lst:extractPreds}@
  m = createSVMModel(preds, posTrainEs, negTrainEs);
  return model;
}

generateDecomposableModel (model) {
  supportVectorEs = getSupportVectors(model);
  decompModel = callSketch(supportVectorEs);
  return decompModel;
}

activeLearningRound (posEs, negEs, unratedEs, @\label{lst:activeLearning}@
                     numSamples) {
  model = learnModel(posEs, negEs); @\label{lst:learnModel2}@ 
  for (e in unratedEs)
    ratings[e] = computeRating(model, e);
  sortedEvts = sortByAbsValue(ratings);
  decompModel = generateDecomposableModel(model); @\label{lst:genModel}@
  return (sortedEs[0:numSamples], decompModel);
}    
\end{lstlisting}
\end{small}
\end{tabular}
\vspace{-0.2in}
\caption{Hybrid Algorithm}
\vspace{-0.2in}
\figlabel{hybridAlgorithm}
\end{figure}

Learning begins by giving the set of positively labeled (i.e., those labeled
as ``interesting''), and negatively labeled events to \C{learnModel} on line
\ref{lst:learnModel1}, which
first invokes the Sketch synthesizer to generate a number of functions
(the number to generate is a parameter to the algorithm).  The functions
are then passed to \C{extractPredicates} on line \ref{lst:extractPreds}, 
which extracts and returns the
set of base predicates from each function (e.g., \C{user = John}).
The base predicates are
then passed to the SVM to generate a model that returns a numerical rating
ranging from -1 to 1.  The model is used in classification of incoming
events (not shown in \figref{hybridAlgorithm}), 
where the incoming event is negatively labeled if the rating is
less than 0, and is positively labeled otherwise.

Then, during each round of active learning,
\C{activeLearningRound} on line
\ref{lst:activeLearning} is called with 
the list of previously rated events, the
list of unrated events to choose from for subsequent user querying, and the
number of events to choose.
It first constructs a model using
\C{learnModel} on line \ref{lst:learnModel2} 
with the list of previously rated events.  Then, for each 
unrated event, it asks the model to compute its (numerical) rating; events are
then sorted according to the absolute values of their ratings, 
and the ones that are closest
to 0 (i.e., the ones that are the most uncertain according to the current model)
are chosen to query the user for labels.  At the same time, 
\C{generateDecomposableModel} on line \ref{lst:genModel} 
is called to create a model representation to drive
subsequent data acquisition.  
In our experiments the time taken to construct models is typically short.

Noisy data might prevent Sketch from generating any candidate
function since the ratings might be contradictory.  The \C{subsample} function
on line \ref{lst:subsample}
is used as a means to remove contradicting inputs prior to model training.  
Even though more sophisticated methods can be used, our experiments have shown
that the simple sampling method is good enough to give reasonable performance in
presence of noise.

In a sense, one can view the hybrid approach as using the Sketch synthesizer as
a feature selection mechanism, and feeding the selected predicates into the SVM 
to build the resulting classifier.  To test that view, we have implemented other
standard feature selection algorithms and provide comparisons in 
\secref{experiments}.

\subsection{Generating Decomposable Models}
\seclabel{hybridModelExplanation}

The output learned by a linear SVM is a model consisting of a linear function
made up by a selected set of predicates, and a list of the all the input
predicates and weights for each of them. The weights for each predicate are
computed using standard methodology 
from the weights the SVM assigns to each input event instance.
Unfortunately such a model does not decompose well 
into per-device filters usable for further data acquisition.
On the other hand, given the input training data,
a program synthesizer is able to generate a classifier that is 
decomposable, but unfortunately 
program synthesizers do not provide any generalization guarantees.
Fortunately, SVM is able to help us in that respect, since it already identifies
the subset of the training data that is used to define the separating 
hyperplane, otherwise known as the support vectors, and in most cases, the number
of support vectors is much smaller than the size of the entire training set,
thanks to the SVM's regularization feature.
Thus, as a post-processing step, we feed the events that are labeled as 
the support vectors
to the synthesizer and ask it to generate a decomposable model.  
Even though the resulting
classifier generated by the synthesizer might not be exactly the same as the one
generated by the SVM (for instance, it might pick predicates that have low weights
as assigned by the SVM, but nonetheless can still classify the incoming events),
in \secref{modelExplanationExp} we present empirical evidence that our approach 
does indeed generate decomposable models that are similar to what the user 
originally has in mind.

\section{Experiments}
\label{s:experiments}

In this section we present our experimental results.  
The overall goal of the experiments is to compare various aspects of the different
learners.  

\subsection{Methods Compared}
We used the LifeJoin platform for experimental purposes. 
LifeJoin collects events from two different event streams.  One of them
is about the location of users, with fields
(user, location, start time, end time), and the other
one about users' activities, with fields (user, activity, start time, 
end time).  As mentioned in \secref{introduction}, we are interested in
composite events where events from the two streams can be combined in different
ways, for instance joining them on the user fields, and part of the learner's
goal is to learn how to combine the two event streams to generate interesting
events.
In the following we use loc$F$ as a shorthand for field $F$ in 
the location event (and similarly for act$F$ for the activity event), 
and duration is shorthand for the length of
the corresponding event (i.e., end time - start time).
To simplify the description we represent users and activities using numbers
rather than actual names.  We use the {\bf unary} features to describe those
that involve only one comparison operation,
such as locUser = 3, and {\bf conjunctive} features for those that involve
multiple comparisons connected with conjunctions, such as 
(locUser = 3 and activity = 4).  In the rest of the section
{\bf full} refers to the 
set of all unary and conjunctive features together.

For the evaluation we implemented
eight different learners, as shown in \figref{learningSchemes}.
The {\bf L1} and {\bf MI} methods are both classical machine learning approaches 
based on an SVM classifier.
Here the L1 approach uses the LASSO algorithm for feature 
selection, followed by a linear SVM for classification.  The MI 
approach performs feature selection by computing the 
mutual information between
each of the features and the output label, and picks the features with the 
highest scores for subsequent classification using a linear SVM.
Both of these methods enumerate the full feature set on the training data
before feature selection.
The {\bf ensemble} learner represents
the program synthesis approach described in \secref{plApproach}, and
{\bf hybrid} 
represents our new hybrid approach described in \secref{hybridApproach}.

{\bf Tree} is the learner created by first 
learning a decision tree using the C4.5 \cite{C4.5}
algorithm using the weka \cite{wekaWebsite} toolkit, and then creating 
features  
by extracting the path(s) from the root node that leads to the leaves that
classify the event as interesting, as in \cite{decisionTreeFeatureSelection}.
In order to avoid degenerate trees, we lowered the support for splitting and 
did not prune the generated tree.  We have also experimented with 
random trees and the results are similar.  The resulting features are then used
to train an SVM for classification.
 
{\bf Full} is an SVM learner that uses no feature selection on the full set of 
conjunctive features as mentioned above,
{\bf unary} is an SVM 
learner that has no conjunctive features, {\bf poly} uses the same set of 
non-conjunctive features as unary except that the features are passed through a
polynomial kernel.  We did not consider other types of kernels such as radial
basis kernel as they combine the input features in a way that does not produce
decomposable models (req 1 from \secref{introduction}).  For the learners that
involve SVMs, we tuned the parameters (e.g., amount of regularization)
using crossfold validation, and we set the degree of the polynomial kernel to
be 6 after trying all kernels of degree 2 to 8.  We applied the
polynomial kernel to other learners (full, L1, 
MI, ensemble, tree) as well, 
but that did not improve the results.

\begin{figure}
\begin{small}
\begin{tabular} {|p{0.45in}|p{1in}|p{0.7in}|p{0.8in}|}
\hline
Learner & Feature Selection & Classification & Active Learning \\
\hline\hline

{\bf full} & none & linear SVM & linear SVM \\

\hline
{\bf unary} & drop all conjunctive features from the full feature set & linear SVM & 
linear SVM \\

\hline
{\bf poly} & same as unary & poly kernel SVM  & poly kernel SVM \\

\hline
{\bf L1} & LASSO on full feature set & linear SVM & linear SVM \\

\hline
{\bf MI} & compute MI on full feature set, 
and pick features with score above preset threshold & linear SVM & linear SVM \\

\hline
{\bf hybrid} & 10 Sketch iterations on the training set to generate
features & linear SVM & linear SVM \\

\hline
{\bf ensemble} & 10 Sketch iterations on training set to generate
features & 10 Sketch iterations on test set and
majority voting & events with the most \# of disagreements among
the base learners \\ 

\hline
{\bf tree} & same as unary, but use decision tree to pick features & 
linear SVM & linear SVM
\\
\hline
\end{tabular}
\caption{Description of the learners used}
\vspace{-0.2in}
\figlabel{learningSchemes}
\end{small}
\end{figure}

\subsection{Experiment Setup}
\seclabel{dataset}

We generated a synthetic data set in which we modeled 5 users,
randomly and uniformly selecting one of 5 location to visit.
Each user remains at the location for a random period of time (ranging
from 1 to 10 hours), and randomly and uniformly
selects one of 5 activities to perform at the location.  This
is meant to model the type of input data that LifeJoin produces.  
The experiments were run on a server
with 32 cores and 30GB of RAM.  We try to execute the experiments in parallel
as much as we can. We choose to evaluate our methods on synthetic data rather
than actual data since no publicly available large data set is available, and 
using synthetic data decouples us from the potential errors in 
data collection or event identification on the phones.
The data set is generated randomly and does not favor or disfavor 
any particular learner.

In addition to a data set, we need a way to generate user interests 
(for labeling training data and to generate ground truth for purposes of 
evaluating the performance
of the different learners.)  To do this,
we manually created 6 different interest functions  
of increasing complexity and used numerical values to represent users, locations, 
and activities, as shown below.  

\begin{asparaenum}
\item locUser = actUser

\item (locUser = 3 $\AND$ locDuration $>$ 1 $\AND$ activity = 0) $\OR$ 
      (locUser = 0 $\AND$ activity = 2)

\item (location = 3 $\AND$ locUser = actUser $\AND$ locDuration $>$ 3) $\OR$
      (location = 2 $\AND$ activity = 1 $\AND$ actDuration $>$ 2) or
      (locUser = 1 $\AND$ locDuration $>$ 4)

\item same as 3. plus disjunct:
      (actUser = 3 $\AND$ activity = 2 $\AND$ actDuration $>$ 1)

\item same as 4. plus disjunct:
      (actUser = 1 $\AND$ actDuration $>$ 1)

\item same as 5. plus disjunct:
      (locUser = actUser $\AND$ actUser = 2 $\AND$ locStartTime - actStartTime $<$ 2)

\end{asparaenum}

Each of the interests above causes different amount of class
imbalance in the input training data.  For instance, 
the first interest function labels about 40\% of the events to be positive, 
whereas the last (most complicated) interest function
labels only about 10\% of the events as positive.  
This is to model how class
distribution can vary drastically among different user interests.
 
For all the experiments we allow Sketch to learn a maximum of 14 different
interests, and allow each interest to consist of a maximum of 7 different
conjuncts.  The numbers were picked from initial sampling of 5 users.
Obviously limiting to 7 conjuncts is more than needed in order to learn the 
predicates listed above, but we used that setting for two reasons.
First, we believe this level of interest complexity is a reasonable approximation
of the maximum complexity of interests a user might have.  
Second, for the experiments below that contain errors in the training
set, limiting the number of interests to be too small could result in 
Sketch not being able to find a satisfying model. 

To generate training data (and validate the performance of our
learners), we labeled data points in our data set using each of these
interest functions, assigning a positive label to the event for a
given interest function if the interest function evaluates to true. 

\subsection{Cross Validation Experiments}
In the first set of experiments, we evaluate the accuracies of the different schemes
using cross validation.  
The goal of this experiment
is to evaluate learner performance in the absence of any performance anomalies the 
active learning methods may introduce.

For each of the predicates we first generated a dataset of 
100 positively and 300 negatively labeled events.  The events are 
uniformly sampled from a domain consisting of 
5 users, 5 locations, and 5 different types of activities.
We ran 10-fold validation on the dataset, where we divide the positive and 
negative events into 10 partitions.
\Figref{cvNoError-accuracy} shows the average accuracies and 
\figref{cvNoError-features} shows the number of features that are actually used
for classification.

\begin{figure}
\centering
\begin{small}
\begin{tabular} {|c|c|c|c|c|c|c|}
\hline
Learner & Pred 1 & Pred 2 & Pred 3 & Pred 4 & Pred 5 & Pred 6 \\
\hline \hline
full      & 100\%  & 100\%  & 85.5\%  & 79.5\%  & 77.5\%  & 81\% \\
unary     & 75\%   & 75\%   & 75\%    & 75\%    & 75\%    & 75\% \\
poly      & 76\%    & 76.5\%  & 76.5\%  & 76\%    & 75\%    & 75\%  \\

L1        & 100\%  & 100\%  & 92.5\%  & 88\%    & 74.5\%  & 81.5\% \\
MI        & 100\%  & 100\%  & 83\%    & 82.5\%  & 78\%    & 82\% \\

tree      & 100\%  & 100\%  & 91.5\%  & 89.5\%  & 91\%    & 81.5\% \\ 

ensemble  & 100\%  & 100\%  & 98\%    & 96.5\%  & 93.5\%  & 92.5\% \\
hybrid    & 100\%  & 100\%  & 97.5\%  & 95\%    & 94\%    & 93.5\% \\

\hline 
\end{tabular}
\end{small}
\vspace{-0.1in}
\caption{Cross validation accuracies on error-free training data}
\vspace{-0.1in}
\figlabel{cvNoError-accuracy}
\end{figure}

\begin{figure}
\centering
\begin{small}
\begin{tabular} {|c|c|c|c|c|c|c|}
\hline
Learner & Pred 1 & Pred 2 & Pred 3 & Pred 4 & Pred 5 & Pred 6 \\
\hline \hline
full      & 43k & 43k & 43k & 43k & 43k & 43k \\
unary     & 344 & 344 & 344 & 344 & 344 & 344 \\
L1        & 61.5 & 118.6 & 32.6 & 265.5 & 482.9 & 604.9 \\
MI        & 6708.5 & 6906.5 & 6990 & 6696.7 & 6430.8 & 6650.8 \\
ensemble  & 26.5 & 40.1 & 50.2 & 45.2 & 40.2 & 46.7 \\
hybrid    & 27.9 & 43.8 & 43.2 & 39.4 & 39.5 & 41.8 \\
\hline 
\end{tabular}
\end{small}
\vspace{-0.1in}
\caption{Cross validation feature set sizes on error-free data (poly and tree
have the same \# as unary)}
\vspace{-0.1in}
\figlabel{cvNoError-features}
\end{figure}

The results show that our hybrid learner has similar accuracy
 as compared to standard machine learning techniques.  At the same time, 
it does not require using the full feature set as in 
the other learners such as LI or MI.  
This is particularly important when comparing the number
of features that are used for classification.  To achieve the same overall
accuracy, the number of features used by the hybrid and ensemble learners
are an order of magnitude
smaller as compared to others, as shown in \figref{cvNoError-features}.

Next, we repeated the same experiment, but this time we introduced an 5\% error
into the training set. Here error refers to the chance that a given event in 
the training set if mislabeled, i.e.,
an event that is labeled as ``interesting'' is reversed to be ``uninteresting''
and vice versa, but the test set is error-free.  
This is to model human error or user interest drifts over time.

The results shown in 
\figref{cvError-accuracy} is similar to the case without errors, except that
the average accuracies of all the learners are lowered, as expected.  It also
took longer for the experiments to complete (about 1-2 hours per fold) 
due to the complexity introduced by the erroneous events.

\begin{figure}
\centering
\begin{small}
\begin{tabular} {|c|c|c|c|c|c|c|}
\hline
Learner & Pred 1 & Pred 2 & Pred 3 & Pred 4 & Pred 5 & Pred 6 \\
\hline \hline
full      & 94\%    & 92\%    & 81.3\%  & 69\%    & 70.2\%  & 75\% \\
unary     & 65\%    & 64\%    & 62\%    & 70\%    & 66.4\%  & 70.2\% \\
poly      & 75.5\%  & 75\%    & 75.5\%  & 74.5\%  & 74\%    & 75\%    \\

L1        & 92.6\%  & 93.7\%  & 84\%    & 82\%    & 80.5\%  & 74.3\% \\
MI        & 93.4\%  & 90\%    & 88\%    & 80\%    & 78.4\%  & 76\% \\

tree      & 90\%   & 90\%   & 85.5\%  & 77.5\%  & 78.5\%  & 74\%  \\

ensemble  & 95\%    & 95\%    & 82.5\%  & 85\%    & 84\%    & 84.7\% \\
hybrid    & 96\%    & 91\%    & 86\%    & 84.2\%  & 83\%    & 86.5\% \\
  
\hline 
\end{tabular}
\end{small}
\vspace{-0.1in}
\caption{Cross validation accuracies on data with 5\% error}
\vspace{-0.2in}
\figlabel{cvError-accuracy}
\end{figure}

\subsection{Active Learning Experiments}
In the next set of experiments we evaluate the learners in the actual usage 
setting, where the user is asked to label a few new data points each time
she visits her newsfeed.
At the end of each round the learner is given the newly rated events along with
the previously rated ones to refine its model about the user.
The goal of the learner is to select the list of events
to present in each round so as to maximize the accuracy of the model,
and to do so with as few rounds as possible.

\subsubsection{Basic Setting}
For evaluation purposes we generate 100 positive and 300 negative events
as a training set to be presented during active learning.  The events 
are generated using the same settings as in the cross validation experiments.
We then generate an additional 10k events and ratings 
(which are not given to the learners) to use as the test set.  
The events in the test set are generated randomly without regards to the 
 ratio of positive and negative events (about 10\% - 40\% of the test
 events are positive, depending on the interest function).
Initially, the learners are given 1 positive and 1 negative event to 
learn an initial model.  Then, during each iteration, the learners
choose 5 events from the training pool to query for their ratings to 
rebuild the model.
We measure the accuracy of 
the model at the end of each round for 20 rounds.  
\Figref{alNoError-error5-error10-accuracy}(a) 
shows the results on predicate 6, averaged over 10 runs.
The results
for the other predicates are similar but the learning rates tend to be higher
for less complex predicates as explained next.

The focus of these results is the learning rate, i.e., 
the rate at which the accuracy increases.  
As the results show, while the learners 
that use classical feature selection mechanisms (L1 and MI)
do have higher learning rates
as compared to those that do not (full and unary), 
our hybrid and ensemble learners have a significantly higher learning rate than 
any of the others, due to the fact that they are able to pick features with
higher predictive power, as discussed in \secref{hybridApproach}.

\Figref{alNoError-featureSizes} shows the number of features that are used
for classification in each round for the learners.  
While they all increase as the number of rounds increases
as expected, 
the growth rate for the hybrid and ensemble learners that use Sketch for
feature selection is much slower than the others.

As a note, we also experimented with skewed data, where the training and testing 
data are biased towards certain users and locations (to model popular events,
and thus the class imbalance is less severe),
along with another experiment where we varied the number of events to add
per round of active learning.  The results are similar to those from the basic
setting.

\subsubsection{Effect of Errors}
Next, we introduce
errors into the training events as in the cross validation
experiments and quantify the effect on accuracies.  
We introduced $K$\% error.
We run two sets of experiments where we introduced 5\% and 10\% error.
\figref{alNoError-error5-error10-accuracy}(b) shows the results running predicate 6,
with 5\% error, and \figref{alNoError-error5-error10-accuracy}(c) shows the results with
10\% error.

The average accuracy of all the learners is lowered versus the
no error case, as expected.  Also as expected, the 10\% error case
is worse than the 5\% error case.  However, in both cases,  the hybrid learner still performs better
than the others.  The number of features used (no shown) exhibits a similar trend as in
the no error case, except that all learners end up using a larger number of
features as a result of the introduction of noise.  This shows the power of our
approach --- by not having an implicit assumption about the class distribution of
the training data, the hybrid learner performs better than those that do.

\subsubsection{Making Use of Extended Labels}
One of the advantages of the hybrid learner over
the ensemble learner is that the SVM in the hybrid learner 
is able to make use of extended labels.  This is because extended labels
simply change the problem from classification to regression, where
instead of a binary label (e.g., ``like'' or ``dislike''),
 the goal is to predict ratings on continuous a scale from -1 to +1. 
In this experiment, we repeat the same
experiment as in the basic setting but with extended labels for events.
For events that are of interest, the label remains as +1 as before.
For those that are not of interest, the label is negative, but its value 
is computed in the following way.  Given the user's interest expressed
as $N$ disjuncts $\vee d_i$, where each $d_i$ is a conjunction of predicates,
then if the event $e$ fails all disjuncts, the
value of its label is computed as $\min ( \#failed(d_i, e) / \#(d_i, e) )$, 
where $\#failed(d_i, e)$ is the number of predicates that $e$ has failed within
$d_i$, and $\#(d_i, e)$ is the total number of conjuncts in $d_i$.  We chose
to pick the minimum since this represents the minimal number of changes in $e$
that would make the user happy.  We present the accuracy results in 
\figref{regressionExp} for running on predicate 6, and they show that 
the learning rate for the hybrid-regression learner 
is faster as compared to the
ensemble and original hybrid-binary learners.   
This makes sense since the regression learner is
able to make use of the extended information that is embedded within the 
``near miss'' cases in selecting better samples during each round of active 
learning. 

\begin{figure}
\includegraphics[width=\columnwidth]{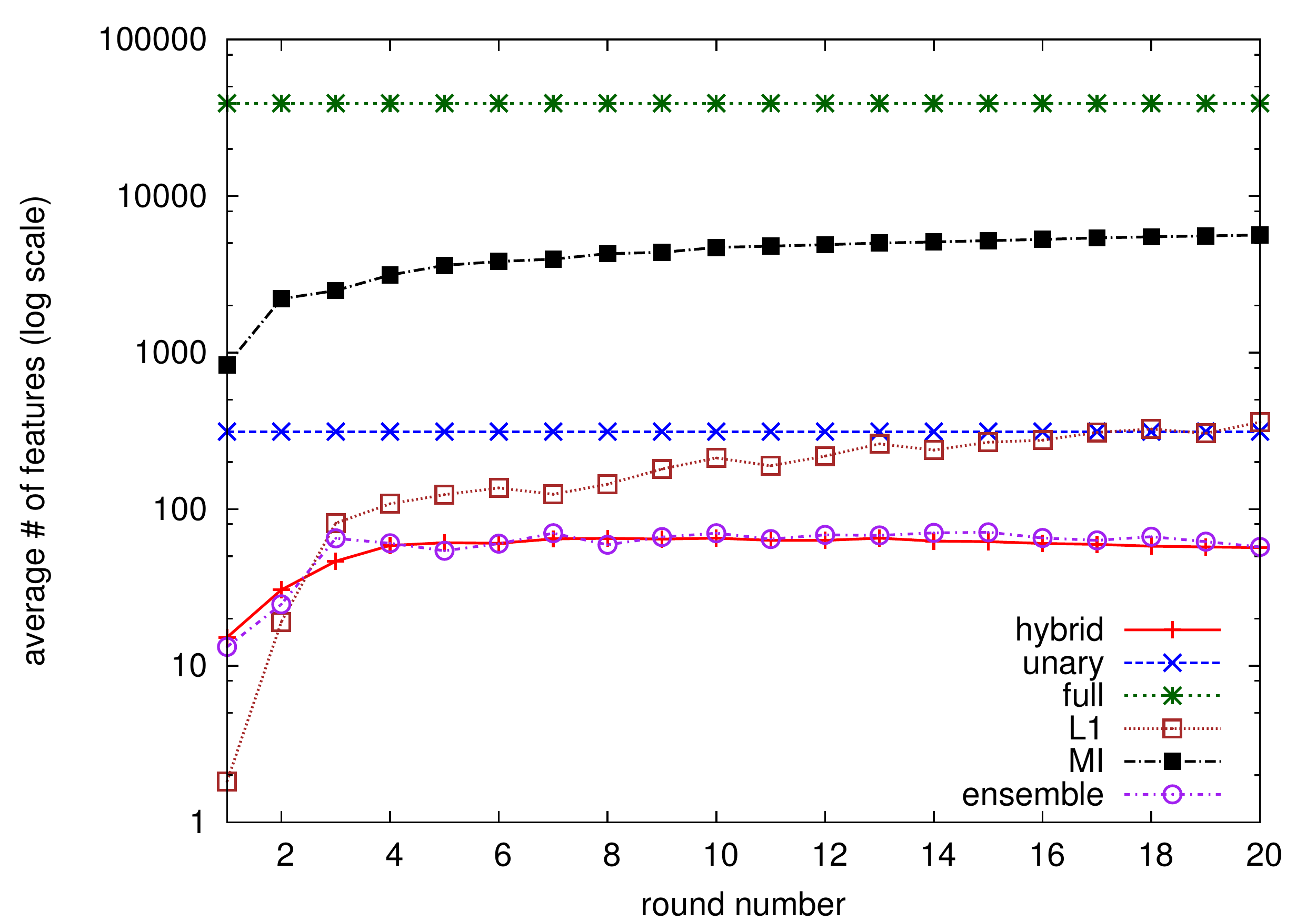}
\vspace{-0.2in}
\caption{Average feature set sizes used by learners on error-free data (poly
and tree have the same \# of features as unary)}
\vspace{-0.2in}
\figlabel{alNoError-featureSizes}
\end{figure}

\begin{figure*}
\hspace*{-.2in}\begin{minipage}{2.5in}
\centering\includegraphics[width=2.5in]{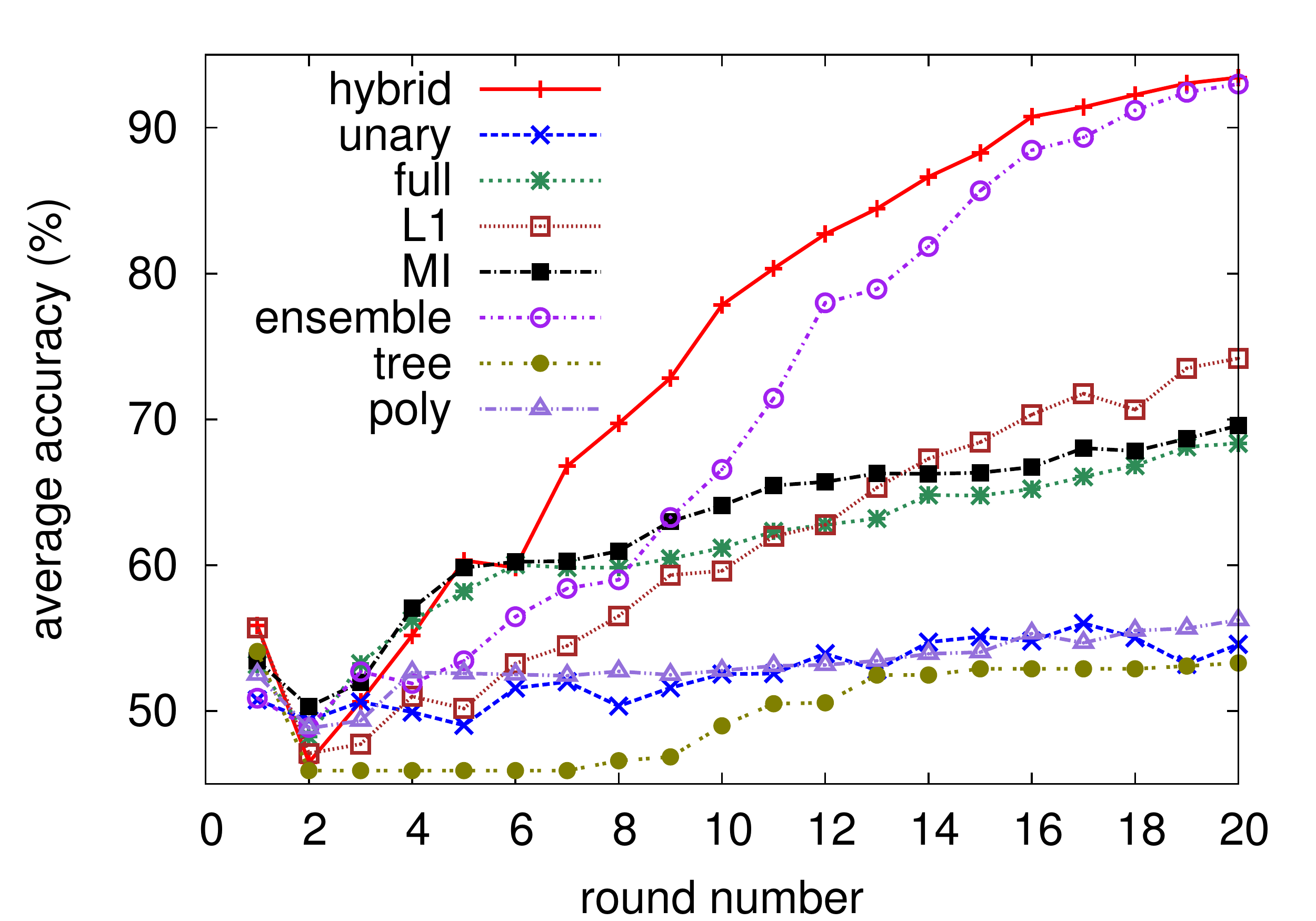}
\centering(a)
\end{minipage}
\begin{minipage}{2.5in}
\centering\includegraphics[width=2.5in]{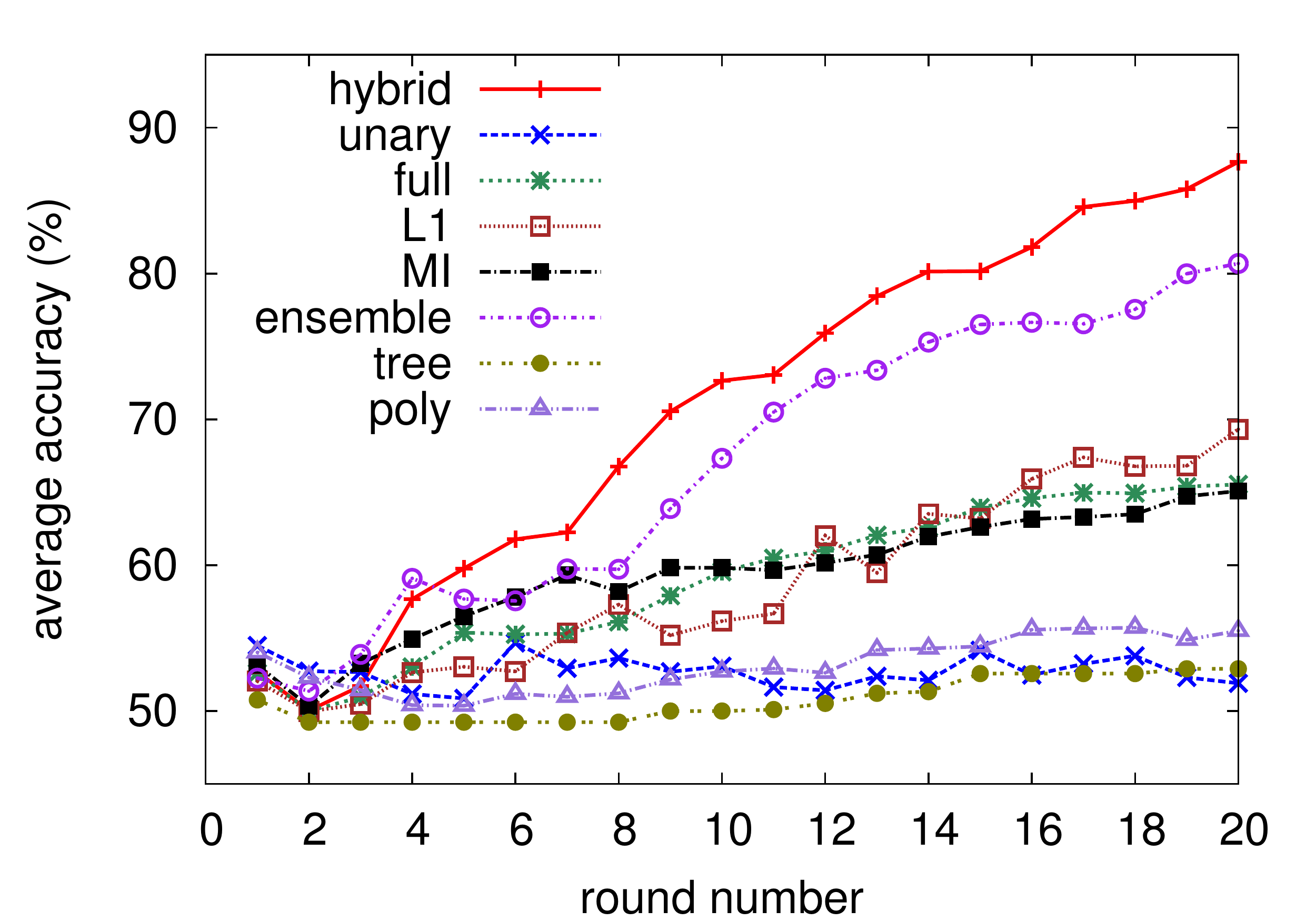}\\
\centering(b)
\end{minipage}
\begin{minipage}{2.5in}
\centering\includegraphics[width=2.5in]{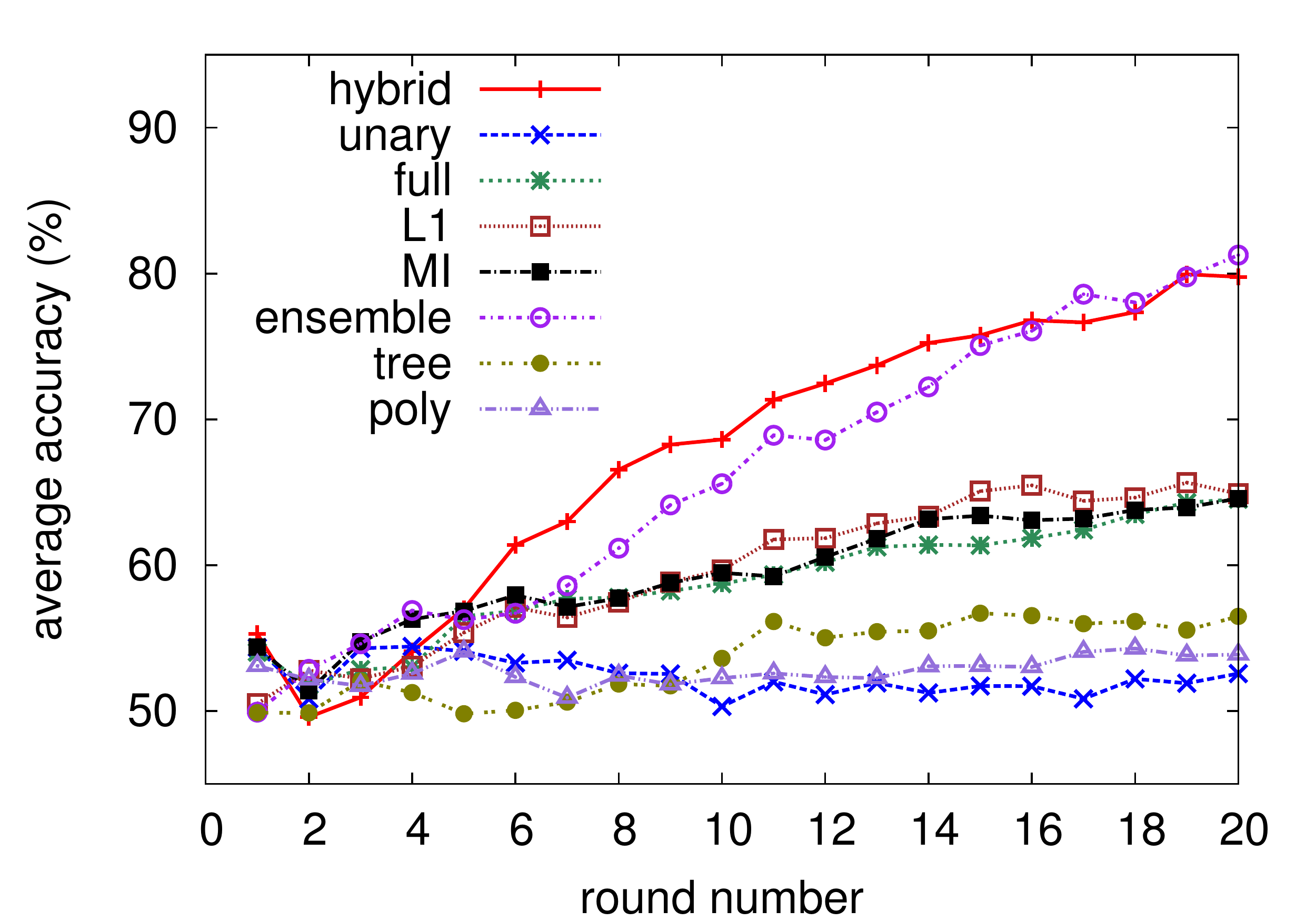}\\
\centering(c)
\end{minipage}
\vspace{-0.1in}
\caption{Average accuracies of learners using (a) error-free, (b) 5\% error, and (c) 10\% 
error data.}
\vspace{-0.2in}
\figlabel{alNoError-error5-error10-accuracy}
\end{figure*}

\begin{figure}
\includegraphics[width=\columnwidth]{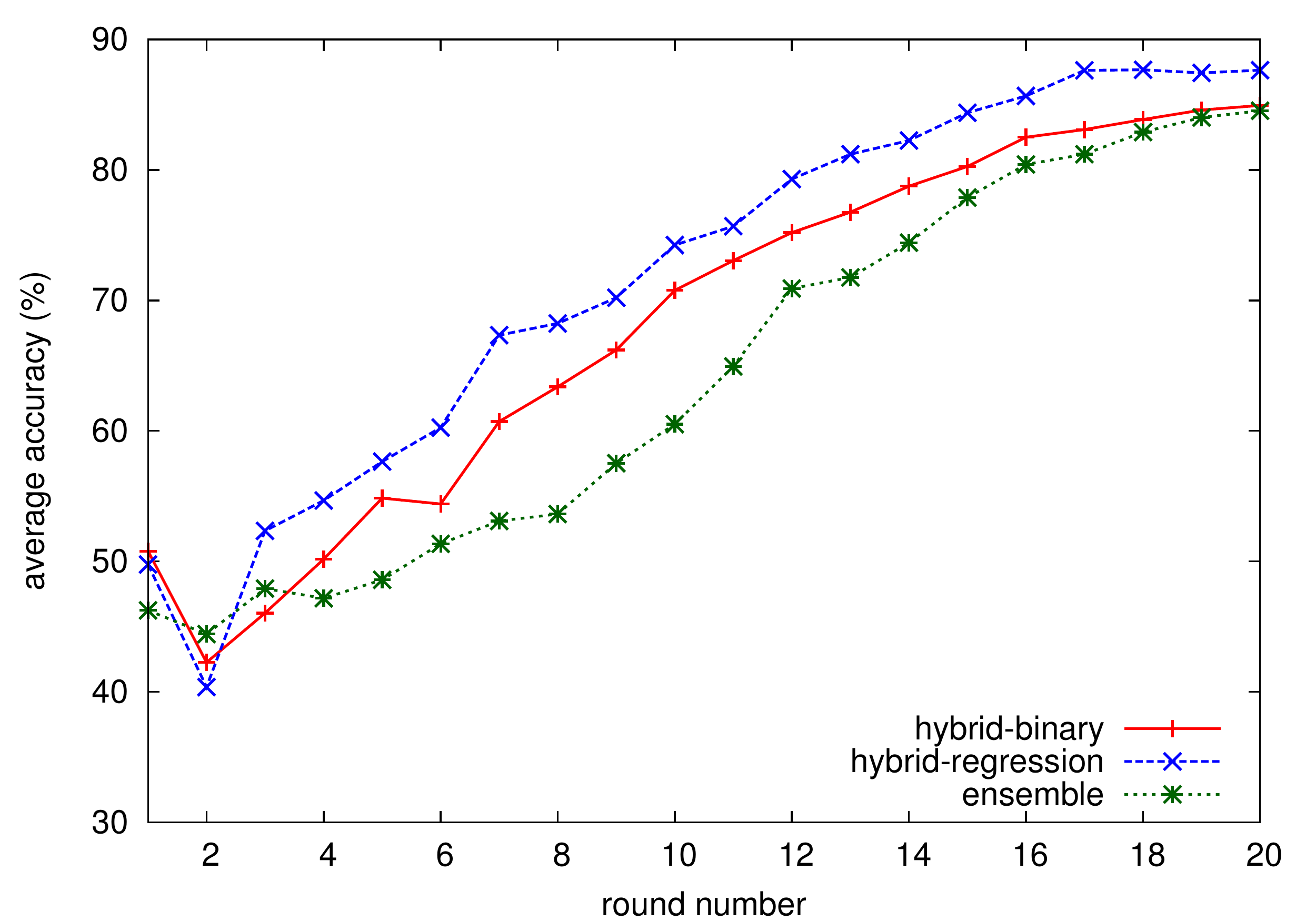}
\vspace{-0.3in}
\caption{Extended label experiment results}
\vspace{-0.2in}
\figlabel{regressionExp}
\end{figure}

\subsubsection{Large Domain}
\label{s:largeDomainExperiment}
In the next experiment we increase the number of users and the number of 
locations from 5 to 50, and the number of activities from 5 to 10.  
This is to model a user
who has more friends and visits more locations.  
We generated the 100 positive and 300 negative training events 
from the new domain using uniform sampling as before, and an additional 
10k events for the test set.
\footnote{\begin{small} Since we are learning a separate model for each user, we 
do not need to scale up to, say 1M for the number of users 
or locations as it is unlikely that a given user would have that many friends or 
locations traveled. \end{small}
} 
We execute the same active learning
experiment as before.
\Figref{alLargeDomain-accuracy} shows the results.

\begin{figure}
\includegraphics[width=\columnwidth]{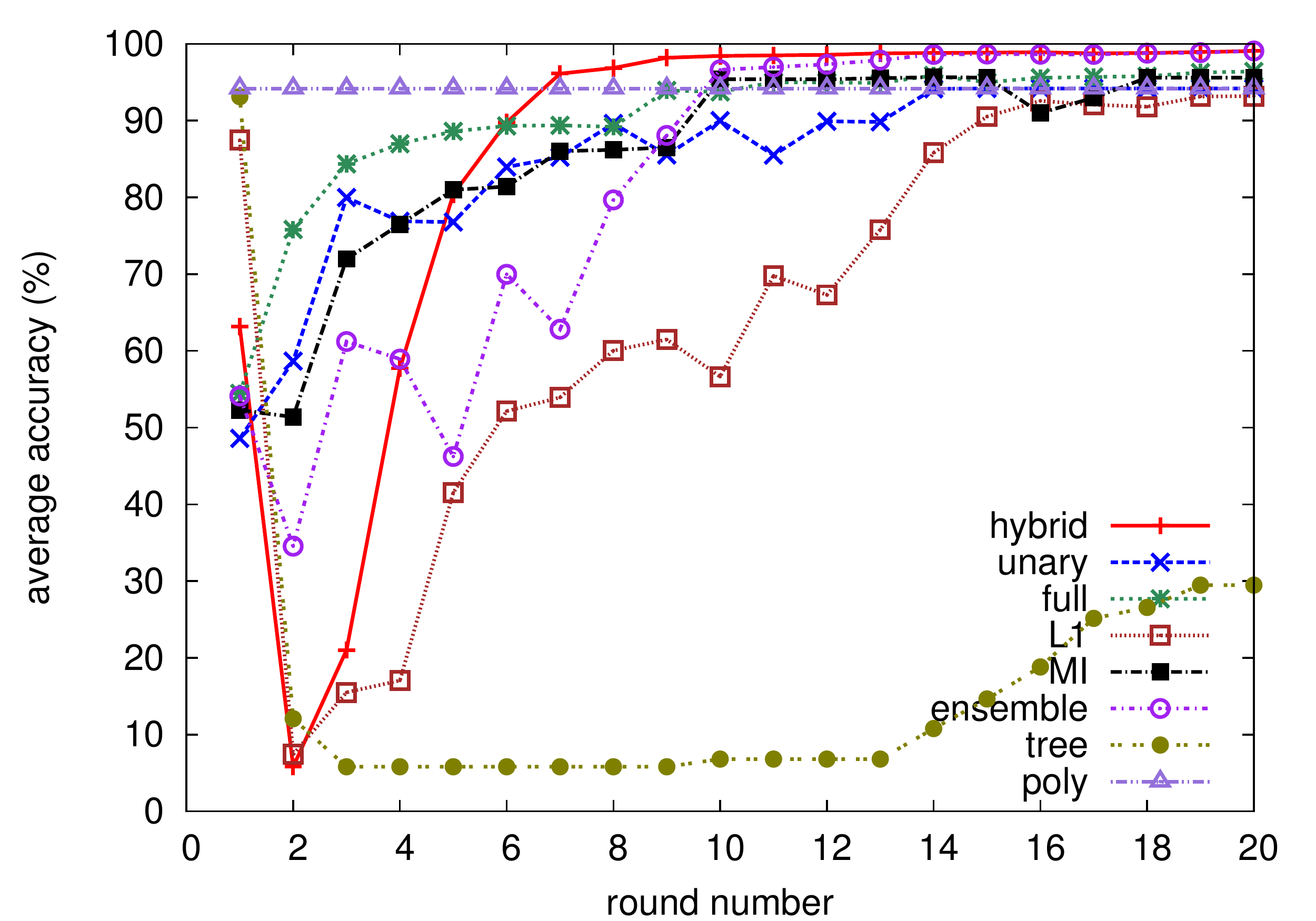}
\vspace{-0.3in}
\caption{Average accuracies of learners using data from large domain}
\vspace{-0.2in}
\figlabel{alLargeDomain-accuracy}
\end{figure}

On the outset, it seems that all learners achieve high 
overall accuracies on the test data,
but close examination proves that not to be the case.  In particular, unlike
previous experiments where the ratio of positively and negatively rated 
events is not heavily skewed, in this case, due to the large domain size,
only around 3\% of the events in the test set are positively rated,  
so the learners quickly learn to assign negative to most test events in order to 
maximize overall accuracy. The result is a model with
high precision on the negatively rated events and very low precision on the 
positively rated ones.  
The decision tree based classifier, however, decided rather to generalize on the
positively labeled events and classifies almost all events as interesting.  As
a result, it achieves high accuracy on the positive events and poorly on the
negative ones, resulting in low overall accuracy.
To illustrate this, \figref{alLargeDomain-posAccuracy} 
show the accuracy results on just the positive events.
The figures show that even though the overall accuracies of the learners
are comparable, the hybrid and ensemble 
approaches actually perform much better than the 
other learners on the positive events.  

\begin{figure}
\includegraphics[height=2.1in, width=\columnwidth]{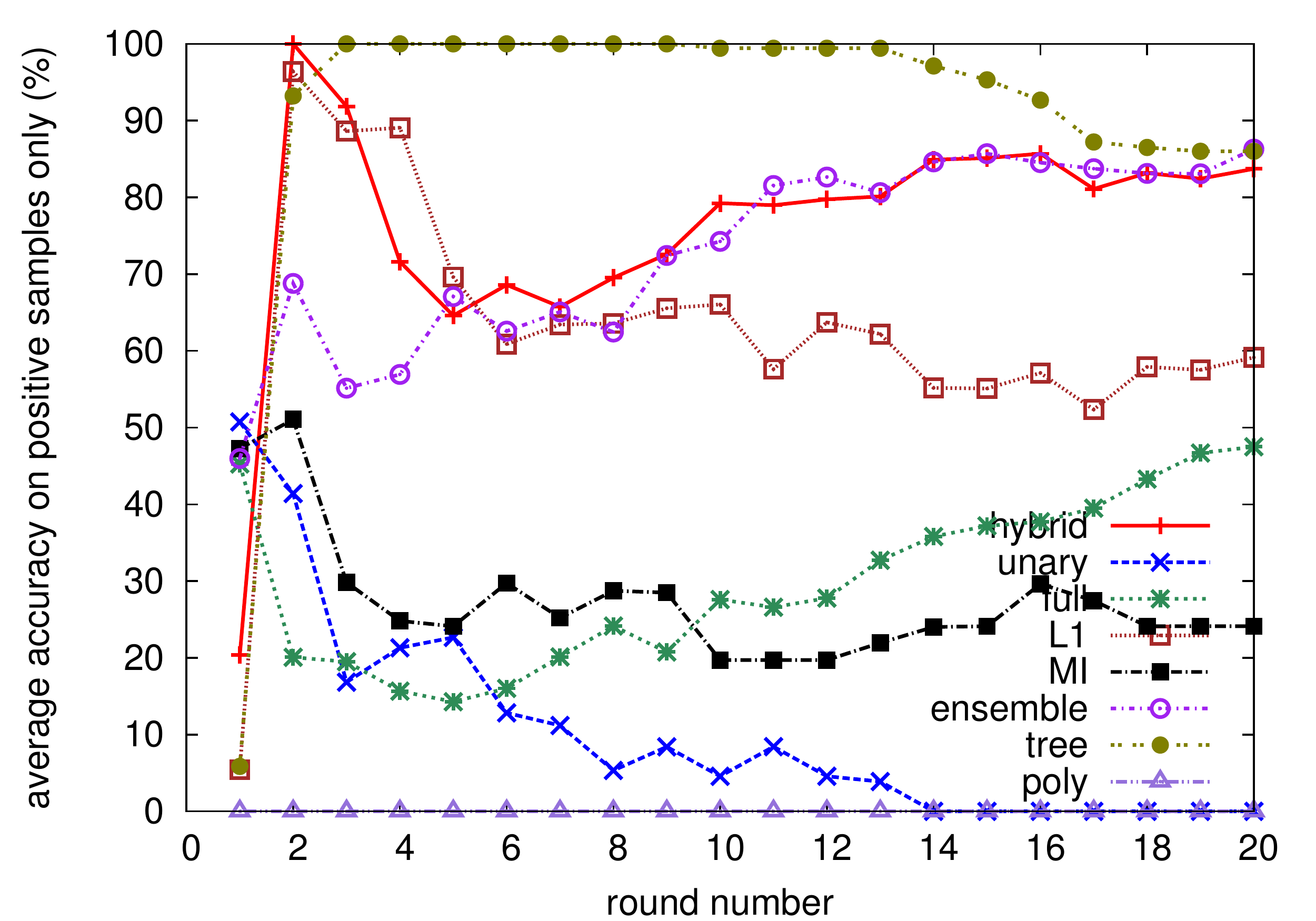}
\vspace{-0.3in}
\caption{Accuracies of learners on positively labeled events}
\vspace{-0.1in}
\figlabel{alLargeDomain-posAccuracy}
\end{figure}

This experiment raises two important points when comparing among 
the learners.  First, all of the learners except for hybrid,
ensemble, poly, and tree 
require enumeration of the full feature set for all events.
In this large domain case, this takes a substantial amount of time 
(2 hours for conversion into the feature representation)
and disk space (300 MB needed to encode 10k events),
as compared to the synthesis-based feature selection approach used in the
hybrid and ensemble learners, which takes much less time  
(10 min to finish the Sketch runs and seconds to convert the chosen features
into feature-space representation)
and negligible disk space (600 kB to encode 10k events).
Secondly, the fact that the classical machine learning based learners assign 
negative labels to most events means that they will very likely not be able to 
identify any interesting events for the user, which is the ultimate goal.

\subsection{Model Explanation Experiments}
\seclabel{modelExplanationExp}
In these experiments we test the effectiveness of using a program synthesizer
at producing a decomposable model (which will provide human readability and the ability to be pushed down onto a phone for data acquisition purposes).  As discussed in 
\secref{hybridModelExplanation}, we took the 
support vectors after model generation and fed them into Sketch.  In an attempt
to generate a minimal description of the model, we ran Sketch iteratively,
 first assuming the user has only 1 interest and asking Sketch to 
generate a description of the model.  If that fails we increase the number of 
interests until Sketch is able to find a description.  
We took the data from one of 
the cross validation experiments without error consisting of 400 events. 
\Figref{modelExplanation}
lists the number of iterations needed for each of the 
predicates to produce the model description, the number of support vectors
used as inputs, along with the actual description generated.  

\begin{figure}
\centering
\begin{small}
\begin{tabular} {|p{0.1in}|p{0.2in}|p{0.2in}|l|}
\hline
int & runs & SV & Interest function learned \\
\hline
1 & 1 & 21 & locUser = actUser \\
2 & 4 & 157 & (locUser = 3 $\AND$ locDuration $\ne$ 4 $\AND$ \\
        & & & locUser = actUser) $\OR$ (activity = 2) \\
3 & 6 & 120 & (locDuration = 4 $\AND$ locUser = actUser) $\OR$ \\
        & & & (location $\ne$ 0) $\OR$ \\ 
        & & & (locUser = 1 $\AND$ locDuration $>$ 4) \\
4 & 6 & 146 & (location $\ne$ 2 $\AND$ locUser = actUser $\AND$ \\
        & & & locDuration = 4) $\OR$ \\
        & & & (location $\ne$ 2 $\AND$ activity = 1) $\OR$ \\
        & & & (locUser = 1 $\AND$ locDuration $>$ 4) $\OR$ \\
        & & & (actUser $\ne$ 1 $\AND$ activity = 2) \\
5 & 10 & 183 & (location $\ne$ 2 $\AND$ locUser = actUser $\AND$ \\
        & & &  locDuration $>$ 2) $\OR$ \\
        & & & (location = 2 $\AND$ activity = 1 $\AND$ actUser $\ne$ 4) $\OR$ \\
        & & & (locUser = 1 $\AND$ locDuration $\ne$ 2) $\OR$  \\
        & & & (actUser $\ne$ 2 $\AND$ activity = 2) \\
6 & 16 & 198 & (location $\ne$ 2 $\AND$ locUser = actUser $\AND$ \\
        & & & locDuration $>$ 2) $\OR$ \\
        & & & (location = 2 $\AND$ activity = 1) $\OR$ \\
        & & & (locUser = 1 $\AND$ locDuration $>$ 4) $\OR$ \\
        & & & (actUser = 3 $\AND$ actDuration $\ne$ 5 $\AND$ locUser $\ne$ 4) $\OR$ \\
        & & & (actUser = 1) \\
\hline
\end{tabular}
\caption{Model explanations using Sketch}
\vspace{-0.2in}
\figlabel{modelExplanation}
\end{small}
\end{figure}

Although the learned predicates do not perfectly match with the predicates used
to generate the labels for the data, they are quite similar, and are relatively
easy to determine what data to subsequently collect on the phones.  
The results also show the power of using 
SVM to reduce the number of input events that are needed to feed into the 
synthesizer, where in the best case (interest 1)
we only need to give 5\% of the original
training events in order to generate a decomposable description that also
happens to perfectly match the original interest function.

\section{Related Work}
\seclabel{relatedWork}

Recently, many probabilistic modeling approaches have been
proposed that can also be applied to the learning problem discussed in this
paper, including Bayesian networks
\cite{bayesianNetworks}, statistical relational learning
\cite{statRelationalLearning}, and probabilistic logic
\cite{probLogic}.  There are also work in building probabilistic models
predicting 
user behavior \cite{tagPrediction, neighborsNetflix, modelingRelationships, 
sideInformation, trustInference}.
However, as with SVMs, models learned using such techniques tend not to generate
decomposable models.

On the other hand, other inductive learning techniques, 
such as inductive logic programming
\cite{foil, progol}, which aim to learn formulas from the training data, can
produce decomposable models.  However, such tools still assume the input data 
to have certain class distribution, and it is unclear how
feature selection can be done for such techniques.

There are many feature selection algorithms that have been
proposed in addition to mutual information and LASSO.  However, our
synthesis-based approach differs from classification techniques
in that most feature selection techniques
 focus on the
statistical properties of the training data, e.g., approximating the
probability distribution of a feature based on the number of data
points in the training set in which it appears, as in the MI metric.  
Such schemes perform
well when fed with a sufficiently large amount of training data, as
evident in our cross validation experiments, but do not do so well in
cases when the training data size is small, as in our active learning scenarios.

In recent years, the programming languages community has been working on 
programming-by-example problems to synthesize different types of programs
\cite{dataStructureSynthesis, oracleGuidedSynthesis, excelSynthesis}.  
Our work
differs from previous tools in that we require a feature selection mechanism
in place in order to provide reasonable results.
The work of Gulwani in \cite{excelSynthesis} proposes querying the user to 
provide differentiating outputs when the synthesizer cannot decide between
multiple programs that satisfy the same input constraints.
Similar ideas appeared 
in \cite{oracleGuidedSynthesis}.  We generalize this
concept and propose the ensemble learning scheme, 
and further show that a hybrid scheme
that combines synthesis-based feature selection with an SVM for classification can
provide excellent performance for social networking applications like LifeJoin.

\section{Conclusions}
\seclabel{conclusions}

In this paper, we presented a learning algorithm that combines the
strengths of classical machine learning techniques with program
synthesis tools, focusing on personalized social recommendation
applications.  
We showed that a hybrid approach, which first uses program synthesis to
generate base learners, followed by breaking down into individual features
and weight assignment with an SVM, 
significantly improves runtime and classification
accuracy.  
Finally, we
showed that using program synthesis on the {\it output} of an SVM can
yield much simpler, and more human readable models, which help users
understand system behavior and can drive subsequent data collection.

The experiments show that the hybrid approach can significantly outperform
traditional classification schemes on synthetic data, but an important next step
is to validate the results on real-world data. Similarly, more
research is needed in analyzing the generalization properties of the
synthesis-based approach. Understanding its theoretical connections
with classical machine learning-based techniques with help develop further
algorithms that leverage the advantages of the two in improving results.

\bibliographystyle{abbrv}
\balance
{ \bibliography{lifejoin}}

\end{document}